\documentclass[10pt,twocolumn,twoside]{IEEEtran}
\usepackage[usenames]{color}
\usepackage[table]{xcolor}
\usepackage{amssymb}
\usepackage{makecell}
\usepackage{multirow}
\usepackage{rotating}
\usepackage{array}
\usepackage{cite}
\usepackage{graphicx}
 \usepackage{psfrag}
\usepackage{subfigure}
\usepackage{url}
\usepackage{amsmath}
\usepackage[ruled]{algorithm2e}

\usepackage[pagebackref=true,breaklinks=true,linkcolor=red,anchorcolor=blue,            citecolor=green,letterpaper=true,colorlinks,bookmarks=false]{hyperref}

\ifCLASSINFOpdf

\else

\fi

\begin{document}

\title{Texture Classification in Extreme Scale Variations using GANet}

% ~\IEEEmembership{Student Member,~IEEE,}
\author{Li~Liu, Jie~Chen, Guoying~Zhao, Paul~Fieguth, Xilin Chen, Matti Pietik\"{a}inen
\IEEEcompsocitemizethanks{\IEEEcompsocthanksitem Li~Liu, Jie~Chen, Guoying~Zhao and Matti Pietik\"{a}inen are with the Center for Machine Vision and Signal Analysis, University of Oulu, 90014 Oulu, Finland.
email: \{li.liu, guoying.zhao, jie.chen, matti.pietikainen\}@oulu.fi
\IEEEcompsocthanksitem P. Fieguth is with the Department of Systems Design Engineering,
University of Waterloo, Waterloo, Ontario, N2L 3G1, Canada.
email: pfieguth@uwaterloo.ca
\IEEEcompsocthanksitem X. Chen is with the Key Lab of Intelligent Information
Processing of CAS, Institute of Computing Technology, CAS, Beijing, 100190,
China.
email:xlchen@ict.ac.cn}\protect\\
}

\markboth{IEEE Transactions on Image Processing}%
{Liu \MakeLowercase{\textit{et al.}}: MRELBP}

\maketitle

\begin{abstract}
Research in texture recognition often concentrates on recognizing textures with intraclass variations such as illumination, rotation, viewpoint and small scale changes. In contrast, in real-world applications a change in scale can have a dramatic impact on texture appearance, to the point of changing completely from one texture category to another. As a result, texture variations due to changes in scale are amongst the hardest to handle. In this work we conduct the first study of classifying textures with extreme variations in scale. To address this issue, we first propose and then reduce scale proposals on the basis of dominant texture patterns. Motivated by the challenges posed by this problem, we propose a new GANet network where we use a Genetic Algorithm to change the units in the hidden layers during network training, in order to promote the learning of more informative semantic texture patterns. Finally, we adopt a FV-CNN (Fisher Vector pooling of a Convolutional Neural Network filter bank) feature encoder for global texture representation.

Because extreme scale variations are not necessarily present in most standard texture databases, to support the proposed extreme-scale aspects of texture understanding we are  developing a new dataset, the \emph{Extreme Scale Variation Textures (ESVaT)}, to test the performance of our framework. It is demonstrated that the proposed framework significantly outperforms gold-standard texture features by more than 10\% on ESVaT. We also test the performance of our proposed approach on the KTHTIPS2b and OS datasets and a further dataset synthetically derived from Forrest, showing superior performance compared to the state of the art.
\end{abstract}


\maketitle
%\IEEEdisplaynotcompsoctitleabstractindextext
%\IEEEpeerreviewmaketitle

\section{Introduction}
\begin {figure}[!t]
\centering
\setlength{\belowcaptionskip}{-0.7cm}
\includegraphics[width=0.45\textwidth]{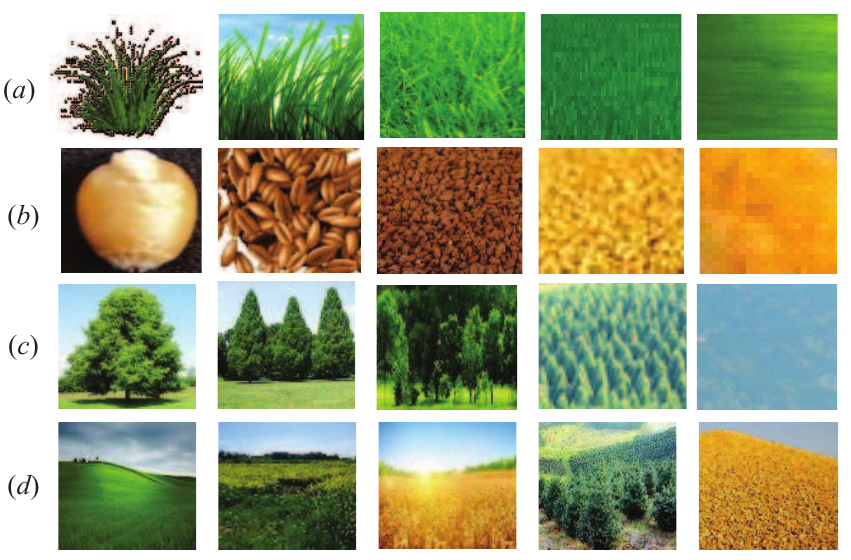}
\caption{A changing scale can have a dramatic impact on the appearance
of textures: (\emph{a}) grasses, (\emph{b}) wheat, and (\emph{c}) trees, with the images from significantly different viewpoints, essentially sampling the textures from a broad extent of underlying continuous scale.  Even more challenging (\emph{d}) are cases of continuous scale variations within a single image.  All images are collected from the Internet.}
\label{Fig:Figure1}
\end {figure}
Texture analysis \cite{Pietikainen11} plays a key role in computer vision, supporting a great many applications: image and scene classification, object detection and recognition, medical image analysis, robot vision and autonomous navigation for unmanned aerial vehicles. Research in texture recognition \cite{Re15Lazebnik05,Pietikainen11,liu2017local} often concentrates on recognizing textures with intraclass variations such as illumination, rotation, viewpoint, and small scale changes. On the other hand, in many real world applications the significant variations or changes of scale may have a dramatic impact on the appearance of an underlying texture, as resolved in some image. For example, as illustrated in Fig.~\ref{Fig:Figure1}, as the scale becomes increasingly and substantially coarser, from left to right, the corresponding texture category also changes; for example, the top row (grasses) changes from grass to lawn to a nearly featureless field.  Particularly in applications such as robot vision or remote surveillance, extreme scale changes can occur quite routinely (Fig.~\ref{Fig:Figure1} (d)), when a single image contains both near and far distances, meaning that in contexts involving autonomous or machine vision it becomes crucial to investigate texture analysis under extreme scale variations.  To the best of our knowledge no existing texture classification methods can handle scale changes of such magnitude. Our aim is to fill this gap and to develop effective methods for classifying textures under the sorts of scale changes illustrated in Fig.~\ref{Fig:Figure1}.

Much effort has been devoted to exploring and developing texture features that are robust to a variety of imaging changes, particularly to illumination, rotation, viewpoint and noise\cite{Re22Ojala02,Re26Shechtman2007,Re5Cimpoi2014,
Re18Liu2016,Re35Wang2016,Liu2016Median}. In terms of change-invariant features, scale variations are amongst the hardest to handle and only modest progress has been made in finding features invariant to even small scale changes\cite{Re15Lazebnik05,Xu10,Re6Cimpoi15}. These methods have demonstrated good performance on benchmark datasets such as Brodatz \cite{Re2Brodatz66}, CUReT \cite{Re7Dana1999}, UIUC \cite{Re15Lazebnik05} and KTHTIPS \cite{Re3Caputo05}, and more recently the OpenSurfaces (OS) dataset \cite{Re1Bell2013}; however all of these datasets exhibit rather modest scale variations, not necessarily representative of the significantly harder problem of texture recognition in the presence of the extreme scale variations of interest here.

Existing successful texture representation paradigms seek to represent textured images statistically as histograms over texels or textons \cite{Re6Cimpoi15,Re15Lazebnik05,LiuFieguthPAMI,Zhang07}. The fundamental question is the {\em scale} of this image, in that  texels may compactly represented when computed at some certain scale, but not at larger and/or smaller scales. The \emph{right} scale is thus part of the definition of the texture and plays an important role, recognized early in the pioneering work of Julesz \cite{Julesz1981}.

Motivated by the above, our method first searches scale proposals and reduces the number of proposals by finding the basic dominant texels that occur most frequently in the image.  Based on the challenges posed by this problem, we then propose a new network GANet and adopt a FV-CNN feature encoder for global texture representation. The main contributions of this work are summarized as follows:
\begin{itemize}
\item To the best of our knowledge, we conduct pioneering investigation towards the problem of recognizing textures exhibiting extreme scale variations, that is, with scale variations of two or more orders of magnitude.

\item When the scale of a texture changes, the category of the resulting texture image also changes at some boundary, boundaries which are important to identify in order to label two images of the same physical material as different texture classes.  This paper offers the first investigation of such scale boundaries.

\item We propose a new network, which we refer to as GANet, which can learn more informative texture patterns by using a genetic algorithm to change the units in the hidden layers during network training.

\item We contribute a large texture dataset consisting of 15,747 texture images having substantial scale variations, in an effort to support the study of texture and scale.
\end{itemize}

\section{Related Work}
Texture can be characterized by statistical distributions of texels or textons, which are defined as repetitive local features that are responsible for the preattentive discrimination of textures \cite{Julesz1981}. The recent literature on texture analysis is vast, and recent surveys can be found in \cite{Re18Liu2016,Re33Xie08,Zhang07}.

The approach in this work is related to the texel size or texture scale. Lindeberg \cite{Re17Lindeberg1998} investigated scale for texture description, suggesting that texture characteristics strongly depend upon it.  Mirmehdi and Petrou \cite{Re21Mirmehdi2000} discussed scale variations in real scenes and used them for the segmentation of color textures. There is recent work focusing on the estimation of the local or global scale of textured images without explicitly extracting texture texels \cite{Re15Lazebnik05,Xu10}.  To search the scale proposals of a given texture image, we adopt the binarized normed gradients (BING) algorithm \cite{Re4Cheng2014bing}, which has been shown to be very efficient and powerful in proposing local salient regions.

Recently, deep Convolutional Neural Networks (CNN) \cite{Re10He2016ResNet,Re28Simonyan2014VGG,Re13AlexNet2012} have demonstrated excellent results in many domains of computer vision, including texture recognition \cite{Re5Cimpoi2014,Re6Cimpoi15,Re18Liu2016,Re27Simonyan2014,Re35Wang2016,Re38Zhoubolei2014}. However, it is a common belief that existing CNN architectures are not robust to appearance variations such as rotation, scale and noise \cite{Re18Liu2016}, and the texture recognition work on CNN mainly focuses on domain transferability \cite{Re5Cimpoi2014,Re6Cimpoi15}.

Motivated by the challenges posed by recognizing textures exhibiting extreme scale variations, we propose a new network, GANet, where we use a genetic algorithm (GA) to change the units in the hidden layers during network training in order to promote the learning of more informative semantic texture patterns and to suppress the number of nonsemantic ones. There certainly has already been work applying GA to deep learning \cite{Re8David2014,Re14Lamos2012,Re29Steininger2016,Xie2017ICCV,Ding2013}: The work in \cite{Xie2017ICCV} aims at learning the architectures of modern CNNs by employing an encoding method to represent each network architecture in a fixed length binary string; in \cite{Re14Lamos2012}, a GA was used to train networks with a large number of layers, each of which was trained independently to reduce the computational burden; in \cite{Re8David2014}, a GA was used to improve the performance of a deep autoencoder and to produce a sparser neural network; and in \cite{Re29Steininger2016} a GA was used to train a network when annotated training data were not available.

Our method is quite different from previous work \cite{Re8David2014,Re14Lamos2012,Re29Steininger2016,Xie2017ICCV,Ding2013}, in that our work aims at developing CNNs specifically to learn more semantic patterns, a focus which is has not been studied.

\section{Methodology}

Our work builds on the extensive literature on CNNs in texture recognition \cite{Re6Cimpoi15}, but further motivated by the work of Zhou {\em et al.} \cite{Re38Zhoubolei2014}, which demonstrate a relationship between semantic units and recognition success.  In particular, those CNN convolutional units showing semantic patterns are regarded as effective at visual recognition, while those showing non-semantic patterns are regarded as being incompletely learned, on which basis the authors claimed that increasing the number of semantic units improves the recognition performance.  As a result, we are inspired to adopt a genetic algorithm (GA) to promote the learning of networks in a global and optimized way, by which we aim to reduce the number of non-semantic units and to increase the number of semantic ones.

GAs are effective at searching large and complex spaces in an intelligent way to approximately solve global optimization problems. Furthermore, from weak learning theory in pattern recognition \cite{Webb2011}, using an ensemble of models boosts classification performance, since multiple models capture richer semantic units than a single model does, thus we propose to adopt three CNN models in our GANet.  We will define genetic operations of mutation and crossover, so that we can traverse the search space efficiently, seeking to maximize the number of semantic patterns, thereby successively eliminating non-semantic ones.  By using different training sets to train the three networks, we realize further improvements in unit diversities of the hidden layers, increasing the string diversities used for crossover and mutation, which we expect should lead to improved performance.

\subsection{Proposed GANet Network}
\label{Sec:GANet}

\begin {figure*}[!t]
\centering
 \includegraphics[width=0.95\textwidth]{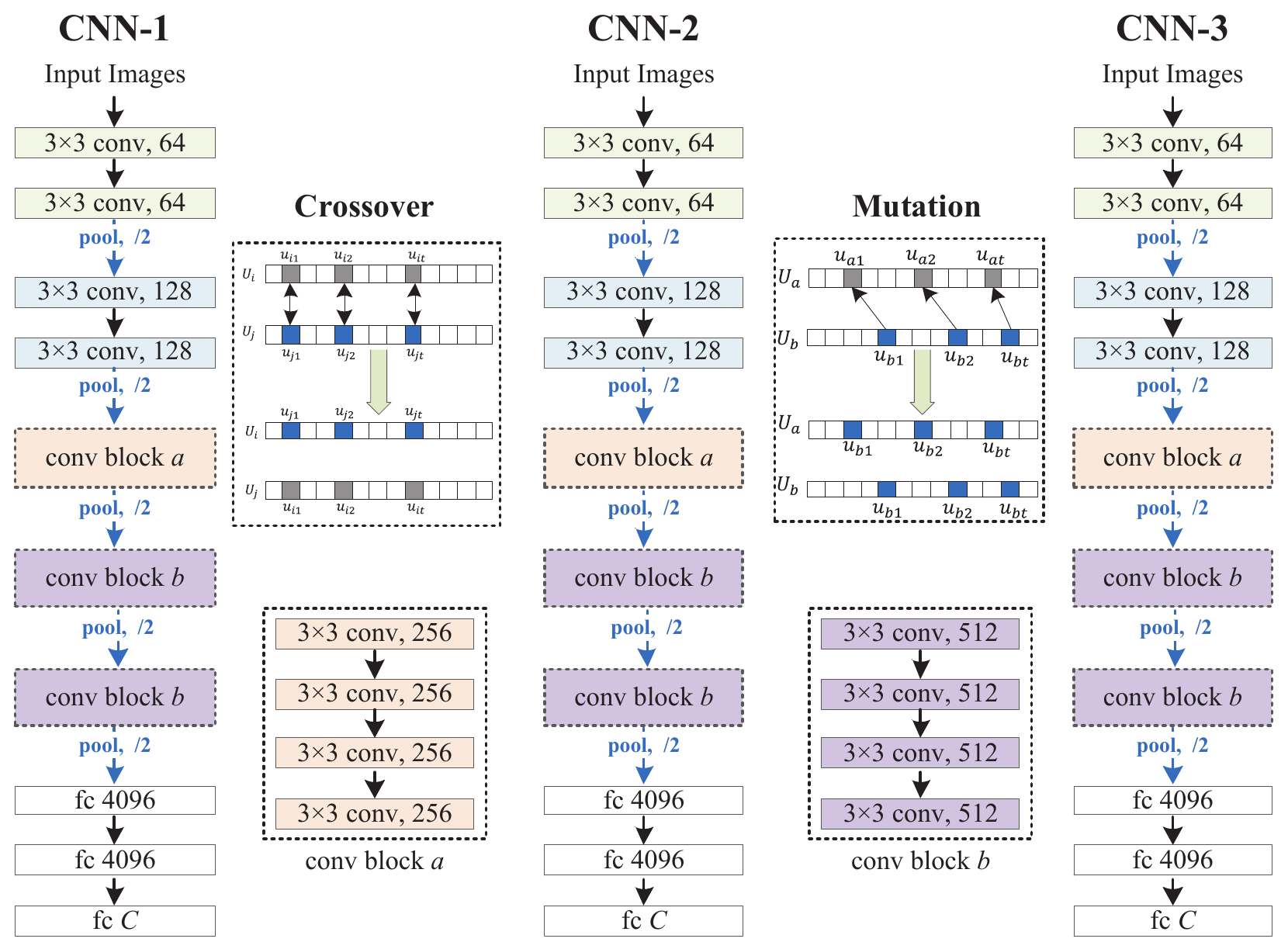}
\caption{Architecture of our GANet network. There are three nets CNN-1, CNN-2 and CNN-3,  each one being a 19 layer VGG-Net. Here, ``$3\times3$ conv 256'' implies a convolution with filters of size $3\times3\times256$, where 256 is the number of feature channels in this layer. ``fc n'' denotes a fully connected layer of size $n$, where $n = C$ (final layer) is the number of texture classes. The GA includes two operations --- crossover and mutation --- which are only applied to the hidden units in the convolutional and pooling layers.}
\label{Fig:GANet}
\end {figure*}

Our proposed GANet is shown in Fig.~\ref{Fig:GANet}, highlighting the genetic operations of mutation and crossover.

{\em Mutation} is a genetic operator used to maintain genetic diversity from one generation to the next of a population of chromosomes, analogous to biological mutation. The mutation process of an individual involves flipping each bit independently with some probability $q$. In practice, $q$ is often small (0.05), since setting $q$ too high causes the search to turn into a primitive random search.  A modest $q$ allows the good properties of a survived individual to more likely be preserved, while still providing opportunities to explore.

In contrast, {\em Crossover} is a genetic operator to vary the programming of a chromosome or chromosomes from one generation to the next. It is analogous to reproduction and biological crossover, involving a swapping, with probability $p$, between two individuals, allowing the diversities of the units in the hidden layers to be improved.

The number of units used for crossover or mutation is not a trivial question, normally found empirically.  Based on our experiments we have choosen 10\% of the units for exchange during crossover and 5\% of the units for mutation.

To undertake the GA operations, all of the units in a convolutional/pooling layer are concatenated into one string. For crossover we choose two unit strings $U_i$ and $U_j$ from two nets at random, and then exchange some elements between these two strings.  For example, as shown in Fig.~\ref{Fig:GANet},
\begin{multline}
\{u_{i1},u_{i2},...,u_{it}\} \text{ from $U_i$ are exchanged with } \\
\{u_{j1},u_{j2},...,u_{jt}\} \text{ from $U_j$.}
\end{multline}
For mutation we similarly choose two unit strings from two nets, however unlike exchange, mutation uses some elements in one string to replace some in the other.  For example, again as shown in Fig.~\ref{Fig:GANet}, we use elements $\{u_{b1},u_{b2},...,u_{bt}\}$ from $U_b$ to replace certain elements in $U_a$, but keep $U_b$ unchanged.  Note that in all cases the two unit strings for crossover and mutation are from the same layer.

The key question is the selection of units for crossover and mutation.  We have developed the following rules:
\begin{itemize}
\item {\em Crossover} elements are chosen at random from the two networks.
\item {\em Mutation} elements are chosen such that, paradoxically, {\em non}-semantic filters replace semantic ones.  Although the targeted removal of semantic filters feels perverse, the approach is, in fact, analogous to the strategy of dropout \cite{Re11hinton2006} in deep network learning, intentionally designed to slow down the learning process, such that the occasional removal of semantic filters leads to more effective overall learning.
\end{itemize}
What remains to be determined is a criterion by which a filter is judged to be semantic or not.  Motivated by the method presented in \cite{Re38Zhoubolei2014}, we visualized the units and found that units at the first convolutional layer are typically responsive to simple texture patterns, such as line elements, crosses and corners, whereas deeper layers are associated with more complex patterns having higher level semantics.

Therefore at the first layer each we apply the Local Binary Pattern (LBP) operator \cite{Re22Ojala02} to the layer, where each location in the layer is considered semantic (systematic, non-random) if its corresponding LBP pattern is uniform, and likewise non-semantic (irregular, random) if the corresponding LBP pattern is nonuniform \cite{Re22Ojala02}.

At coarser layers, the distinction is a bit more subtle.  We begin with a pre-trained VGGNet, from which (as in \cite{Re38Zhoubolei2014}) we visualize the units of each layer and then cluster the units into two groups (semantic and nonsemantic). The minimal activation of the semantic group is used as the semantic threshold $\tau_l$, which will be a function of layer $l$.  Then, for some input image $I$, let $f_j^l$ be the output of the $j$th convolutional filter at the $l$th layer, that is, such that $f_j^l$ includes the effect of the activation function (here a ReLU --- Rectified Linear Unit).  A given filter output at position $(r,c)$ and layer $l$ will be considered semantic if $f_j^l(r,c) > \tau_l$.

For each position $(r,c)$ in feature map $x_j^l$, we wish to determine whether the region around $(r,c)$ corresponds to semantic behavior.  To do this, we search within a $5 \times 5$ window centered at $(r,c)$ to count the number of semantic activations, as just defined.  If we cannot find $k$ (typically $k=10$) strong activations, we neglect position $(r,c)$ and move to the next position in the feature map.  If we do have at least $k$ activations, we assert filter $j$ to be semantically meaningful, an assessment to be taken into account during mutation and in the final assessment of the resulting network.

To learn our semantically driven network, we start with an off the shelf CNN model (VGGNet \cite{Re28Simonyan2014VGG}) pretrained on a large scale dataset like ImageNet \cite{Re27Simonyan2014}.  The model is then fine tuned via the genetic strategy of Fig.~\ref{Fig:GANet}, given texture images.  Based on our experiments, we found the fraction of units showing semantic patterns increasing from 60\% in the original VGGNet (fine tuned on textures, but with no GA) to around 70\% in GANet (VGGNet after the genetic algorithm).

\begin {figure}[!t]
\centering
\includegraphics[width=0.4\textwidth]{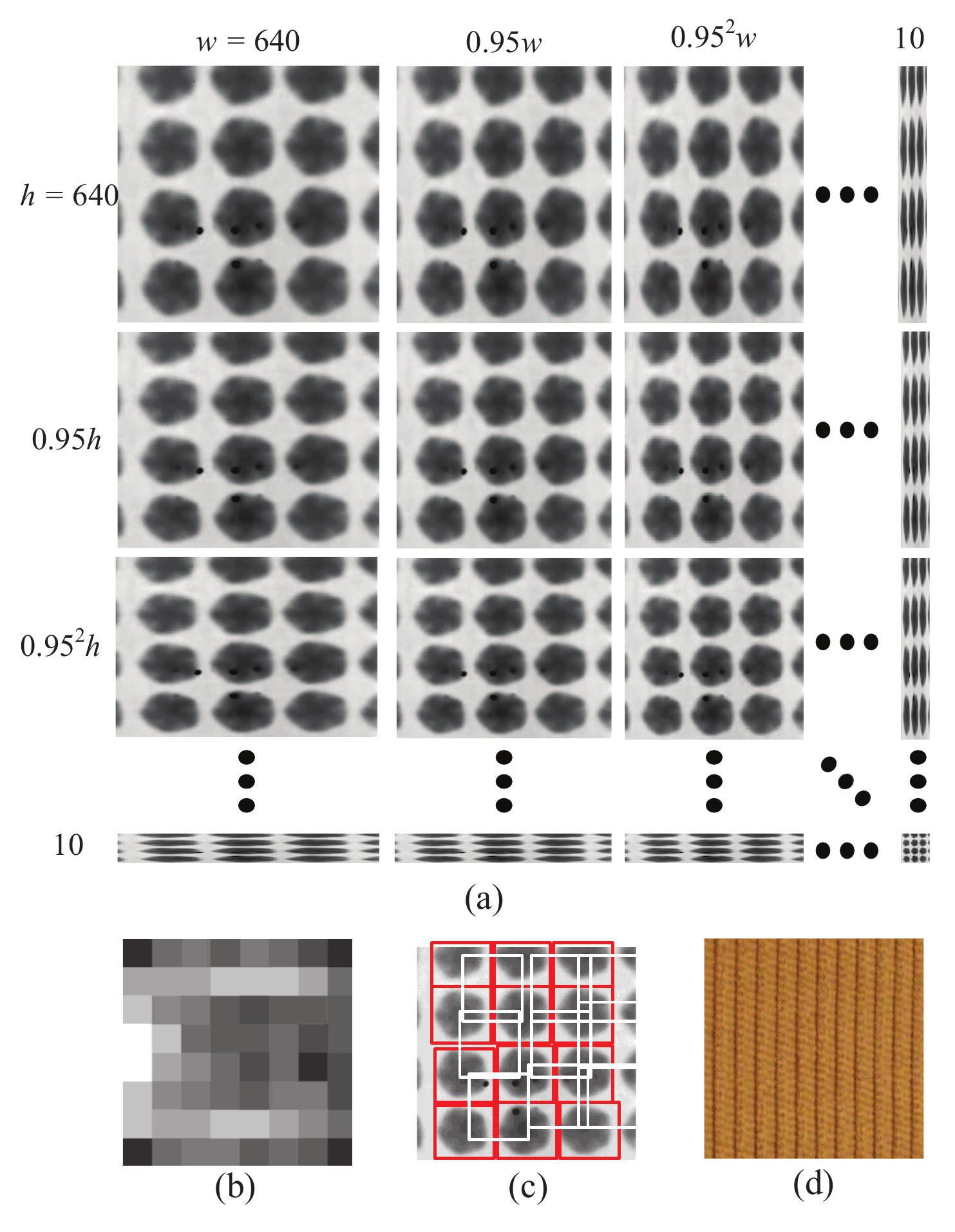}
\caption{Scale proposal searching using an image pyramid and the BING searcher \cite{Re4Cheng2014bing}: (a) Images are downsized to obtain an image pyramid; (b) A single 64D linear model for selecting texture element proposals based on normed-gradient features; (c) Candidate texture element windows; (d) A texture image whose texels have a significantly different aspect ratio from those in (c).}
\label{Fig:Figure3}
\end {figure}

\subsection{Scale Proposal}
\label{Sec:ScalePro}
Intuitively, the scale change of a texture image clearly affects its appearance, however the direct estimation of global scale is very unstable. In order to achieve scale invariance in texture classification we propose to search a set of scale proposals or candidate scale levels in a given texture image.  We then reduce the number of scale proposals by searching among the basic dominant texels that occur most frequently.

\textbf{Scale Proposal Searching.}
In general, there can be not only multiple scales, but indeed {\em continuous} changes in scale within a single image, as was illustrated in Fig.~\ref{Fig:Figure1} (\emph{d}).  A necessary prerequisite for texture classification with extreme scale variations to be successful is that we should find all of the existing scale proposals; to this end, we use the BING algorithm \cite{Re4Cheng2014bing} to find candidate texture element (texel) windows, and then compute the scale proposals according to these windows. The rationale behind the BING searcher is that it searches the scale proposals in real time and returns almost all of the potential texels in an image.

As demonstrated in Fig.~\ref{Fig:Figure3} (\emph{a}), to be sensitive to a range of scales  we resize an input image $\textbf{I}\in\mathbb{R}^{W_0\times H_0}$ to a sequence of quantized sizes characterized by scale ratio $s$. In our experiments, we choose $s=0.95$ and generate an image pyramid of resized images of sizes $\{(W_0s^{m}, H_0s^n)\}$, to some lower limit (here set to ten pixels), determined by the region of support of the features being extracted, with an $8\times8$ feature extraction window recommended by Cheng \emph{et al.} \cite{Re4Cheng2014bing}.  We calculate the \emph{normed gradient} (NG) \footnote{The normed gradient represents Euclidean norm of the gradient.} feature \cite{Re4Cheng2014bing} (shown in Fig.~\ref{Fig:Figure3} (\emph{b})) of the entire pyramid.  Note that we deliberately downsample separately along each axis, to account for textures having different aspect ratios (as illustrated in Fig.~\ref{Fig:Figure3} (\emph{d})).

To find texels within a texture image, we scan over its entire image pyramid with an $8\times8$ BING feature \cite{Re4Cheng2014bing}.  As shown in Fig.~\ref{Fig:Figure3} (\emph{c}), at any particular scale level $l$ a number of texels $\Omega=\{T_{l,k}\}$, indexed by location $k$, could be proposed.  To keep the concept of scale clear, every location in the image pyramid is characterized by its rescaling relative to the original image; that is, pyramid image of size $\{(W_0s^{m}, H_0s^n)\}$ is said to be at scale $l=(s^m, s^n)$.  We denote all texels found over the entire image pyramid as $\Omega = \{T_{l,k}\}$.

In practice, we typically find thousands of potential texels over an entire image pyramid by using small thresholds for the BING searcher. The scales of these texels form the scale proposals for this texture image. Specifically, we denote the set of all the scale proposals
\begin{equation}
\mathbb{S}_{sp}=\{l|T_{l,k}\in\Omega\}
\end{equation}
as the candidate scale proposal.

\textbf{Scale Proposal Reduction.}
The candidate scale proposals $\mathbb{S}_{\textrm{sp}}$ are usually redundant for an input image. For efficiency consideration, we need to reduce the number of the scale proposals by finding the basic texels that most frequently occur in the image.

Textures, whether they are regular or stochastic, contain
repetitive patterns that exhibit stationary statistics of some sort \cite{Re33Xie08}. Hence, the texels found by the BING searcher are expected to appear repetitively. For each texel proposal $T\in\Omega$, we search its similar texels over the texel proposal set from the same scale level. We only keep those texels having sufficiently many similar texels and remove the others from the candidate texel set $\Omega$, with the remaining texels forming a new set $\Omega^{\textrm{re}}$, having a corresponding reduced scale-proposal set $\mathbb{S}_{\textrm{sp}\textrm{-}\textrm{re}}$, a significant reduction in the number of scale proposals.

Based on our analysis, in order to obtain a good reduced scale proposal set, two more issues have to be taken into consideration:  how to evaluate the similarity between two texels, and how many texels to keep.

Regarding the similarity measure between two texels, we propose to compute the distance between the $\textrm{LBP}_{8,1}^{u2}$ histograms \cite{Re22Ojala02} of them.

In terms of which texels in the candidate texel proposal set should survive, for each $T\in\Omega$, we find the set $\Omega^\prime$ of similar texels whose similarities are larger than some threshold $\eta$, such that candidate $T$ is preserved only if the number of similar texels surpasses threshold $K$, only keeping the dominant, most frequently occurring texels, essentially those which are more stable and removing noise.  The setting of the two threshold parameters $\eta, K$ will be discussed in Section \ref{Sec:Results}.

\subsection{Scale Boundary}
\label{Sec:ScaleBoundary}
As demonstrated in Fig.~\ref{Fig:Figure1}, we argue that the category of a texture image only remains unchanged during some scale interval. In other words, when the scale of a texture image changes significantly, the category of this texture image may also change.  The interesting question is the location of the {\em scale boundary} which separates the  two images of the same physical material as different texture classes.  Olshausen and Field \cite{Re23Olshausen1996} reconstruct any given image in a sparse way based on a selected group of patches.  Inspired by this finding, we similarly develop a patch based method to infer boundary in scale.

\begin {figure*}[!t]
\centering
\includegraphics[width=0.98\textwidth]{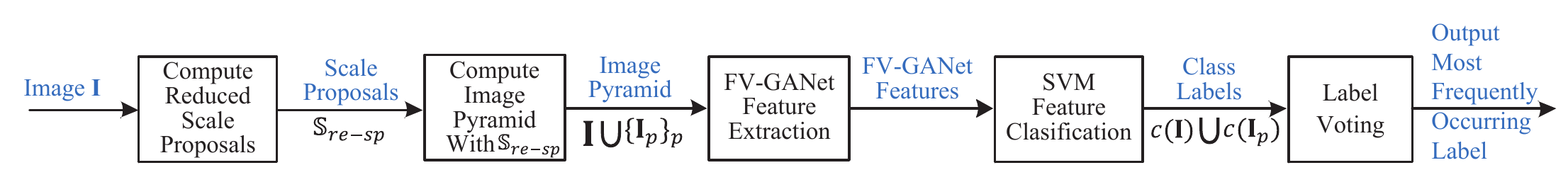}
\caption{The proposed texture classification pipeline}
\label{Fig:Pipeline}
\end {figure*}

Given a texture dataset $\mathbb{S}=\{\mathbb{S}_c\}$ with $C$ classes and for each class $\mathbb{S}_c$, we randomly select 10 images and then compute its basis functions $\mathbb{X}_c=\{x_c\}$ on $16\times16$-pixel patches, as in \cite{Re23Olshausen1996}.  For any image $\textbf{I}\in\mathbb{S}_c$, we compute its reconstructed image $\hat{\textbf{I}}$ and the  reconstructed error
\begin{equation}
\delta=\|\textbf{I}-\hat{\textbf{I}}\|/\|\textbf{I}\|
\end{equation}
If $\delta>\xi$, we consider that category of the texture image changes, \emph{ i.e.} $\textbf{I}\notin\mathbb{S}_c$. In our case, we set $\xi=0.1$ as suggested by \cite{Re23Olshausen1996}.

We adopt this approach to group our synthesized image set SForrest $\Theta^+(\mathbb{S}_f)$  and ESVaT (detailed later in Section \ref{Sec:Data}) into subcategories. The process is detailed in Algorithm 1, where dataset SForrest is used as example.  After applying Algorithm 1, each texture class in SForrest is regrouped into a number of subcategories, \emph{i.e.} $\Theta^+(\mathbb{S}_f)=\{\mathbb{S}^{syn}_{c,p}\}$, indexed by class $c$ and scale level $p$, where there are $P_c$ distinct scale levels associated with class $c$.  As a result, dataset SForrest with $C$ original texture classes has been regrouped into $\sum_{c=0}^{C-1}{P_c}$ categories \footnote{In fact we manually combine some subsets in $\{\{\mathbb{S}^{syn}_{c,p}\}_p\}_c$ as one category since some of them are basically textureless and only show a region of a certain color.}. All of the images are also manually checked after this automatic regrouping. Likewise, ESVaT is also regrouped with Algorithm 1.

\subsection{Summary of Proposed Framework}
\label{Sec:Pipeline}
Our texture classification pipeline is summarized in Fig.~\ref{Fig:Pipeline}, consisting of the following steps:
\begin{enumerate}
\item For each image $\textbf{I}$ in training/testing set $\Theta_{\textbf{I}}$, we compute its reduced scale proposals $\mathbb{S}_{\textrm{sp}\textrm{-}\textrm{re}}$ as described in Section \ref{Sec:ScalePro}.

\item We downsize each image $\textbf{I}$ to obtain its image pyramid $\{\textbf{I}_{p}\}$ using its reduced scale proposals $\mathbb{S}_{\textrm{sp}\textrm{-}\textrm{re}}$. For the downsized images, we let $\mathbb{I}=\textbf{I}\bigcup\{\textbf{I}_{p}\}$ be the expanded set for image $\textbf{I}$. Thus, we have a new training/testing set $\Theta_{\textbf{I}}^+=\{\mathbb{I}\}$.

\item For each image in the new training/testing set, we use the proposed FV-GANet to extract global texture feature representation, following FV-CNN \cite{Re6Cimpoi15}.

\item The extracted FV-GANet features are classified using a SVM classifier. For each image $\textbf{I}$, if its reduced scale proposals $\mathbb{S}_{\textrm{sp}\textrm{-}\textrm{re}}$ has $M$ scale levels (\emph{i.e.} $|\mathbb{S}_{\textrm{sp}\textrm{-}\textrm{re}}|=M$), then we have $M+1$ category labels for it after classification. Image $\textbf{I}$ will be assigned the category label which occurs the most frequently.
\end{enumerate}
The evaluation of the texture classification performance in step 3 uses FV-CNN \cite{Re6Cimpoi15} for texture feature description. FV-CNN truncates a CNN network and regards the last convolutional layer of a CNN as a filter bank, performing orderless pooling of CNN descriptors using the Fisher Vector, as is commonly done in standard bag of words approaches. FV pools local features densely within the described regions removing global spatial information, and is therefore more apt at describing textures than objects. The pooled convolutional features are extracted immediately after the last linear filtering operator and are not normalized. These features are pooled into a FV representation with 64 Gaussian components. FV-CNN is remarkably flexible and effective. First, the convolutional layers behaving like non-linear filter banks are better local texture descriptors than the fully connected layers, which may be useful for representing the overall shape of an object. Second, the FV pooling encoder is suitable for texture description since it is orderless and multi-scale. Third, it avoids expensive resizing of input images since any image size can be processed by convolutional layers.

We have to emphasize that the training set is divided into subcategories via the scale boundary search algorithm presented in Algorithm 1. During classification, a testing image is considered being correctly classified only it is assigned the correct subcategory label.

\begin{algorithm}[t]
%  \SetLine
  \KwIn{The synthesized dataset SForrest $\Theta^+(\mathbb{S}_f)$}
  \KwOut{Regrouped SForrest $\Theta^+(\mathbb{S}_f)=\{\{\mathbb{S}^{syn}_{c,p}\}_p\}_c$}

   \For{each texture class c in SForrest}{
   Let $\Theta^+_c$ be all the samples of class $c$ in  SForrest;

   Set $p=0$;

   \If{$\Theta^+_c\neq\varnothing$}
   {
    1. Randomly select ten texture samples with the \emph{largest} scale level $\mathbb{I}^{\textrm{se}}_c\subset\Theta^+_c$;

    2. Compute basis functions $\mathbb{X}_{c,p}$ based on $\mathbb{I}^{\textrm{se}}_c$;

    3. Reconstruct all the samples in $\Theta_c^+$ with basis functions $\mathbb{X}_{c,p}$;

    4. Determine the samples whose reconstructed error is less than $\xi$ and denote them as $\mathbb{S}^{syn}_{c,p}$: $\mathbb{S}^{syn}_{c,p}=\{\textbf{I}_k|\frac{\|\textbf{I}_k-\hat{\textbf{I}}_k\|}{\|\textbf{I}_k\|}<\xi,\textbf{I}_k\in\Theta_c^+\}$;

    5. Update $\Theta^+_c=\Theta^+_c-\mathbb{S}^{syn}_{c,p}$;

    6. $p=p+1$.
    }
    }
  \caption{Divide each texture class in the synthesized SForrest into subcategories to find its scale boundary}
\end{algorithm}
\section{Datasets and Experimental Setup}
\label{Sec:Data}
\begin {figure}[!t]
\centering
\includegraphics[width=0.45\textwidth]{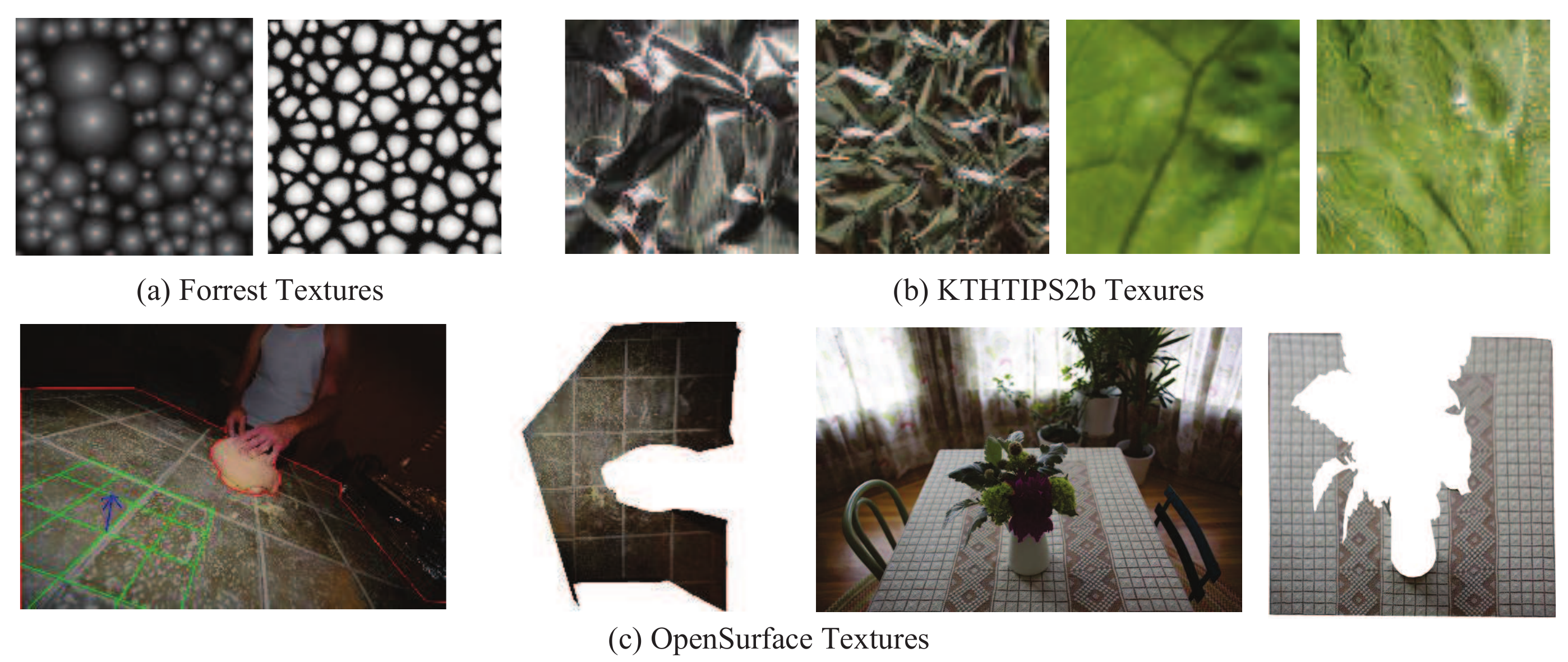}
\caption{Some example textures from (a) Forrest, (b) KTHTIPS2b and (c) OS (Original images and their corresponding annotated texture segments).}
\label{Fig:Figure4}
\end {figure}

We test the proposed framework on four datasets: the synthesized dataset SForrest derived from Forrest \cite{Re39Forrest}, ESVaT, KTHTIPS2b \cite{Re3Caputo05} and OS (OS) \cite{Re1Bell2013}. Some example texture images from Forrest, KTHTIPS2b and OS are shown in Fig.~\ref{Fig:Figure4} and some examples from ESVaT are shown in Fig.~\ref{Fig:Figure1}.

\textbf{SForrest} is synthesized based on the Forrest dataset $\mathbb{S}_f$, which contains 17 texture classes and 935 images captured in the wild. The method for synthesizing SForrest is as follows. For each image $\textbf{I}_i\in\mathbb{S}_f$, we firstly generate an image pyramid $\Theta_{\textbf{I}_i}=\{\textbf{I}_{i,p}\}$ with scale $s=0.95$ using the method detailed at the beginning of Section \ref{Sec:ScalePro} and illustrated in Fig.~\ref{Fig:Figure3} (\emph{a}). We synthesize new images $\Theta_{\textbf{J}_i}=\{\textbf{J}_{i,p}\}$ based on $\Theta_{\textbf{I}_i}$, in that for each $\textbf{I}_{i,p}\in\Theta_{\textbf{I}_i}$, we stitch several reduplications of $\textbf{I}_{i,p}$ together to generate a larger image $\textbf{J}_{i,p}$, which is cropped at random, if needed, to have the same size as $\textbf{I}_{i}$.  We define $\Theta(\mathbb{S}_f)=\{\Theta_{\textbf{I}_i}\}_{i}$ to be the image pyramids of all images in image set $\mathbb{S}_f$, and $\Theta^+(\mathbb{S}_f)$ to be the combined image set $\mathbb{S}_f\bigcup\{\Theta_{\textbf{J}_i}\}_{i}$. $\Theta^+(\mathbb{S}_f)$ is our final synthesized dataset.  For image editing, we use the method proposed by Perez \emph{et al.} \cite{Re25Perez2003}, introduced for the seamless editing of image regions.

\textbf{ESVaT} is composed of 15,747 texture images from 15 material categories\footnote{bark, bubble, brick, carpet, concrete, fabric, grass, granite, laminate, plastic, stone, tile, wood, wheat and tree}, each of which has extreme scale variations and is further annotated to several subcategories by the approach detailed in Section \ref{Sec:ScaleBoundary}. \textbf{KTHTIPS2b} \cite{Re3Caputo05} has 11 texture categories and four physical samples per category. Each physical sample is imaged with 3 viewing angles, 4 illuminants and 9 different scales to obtain different images. From \textbf{OS} \cite{Re1Bell2013} we use the same dataset as in \cite{Re6Cimpoi15}. It has 53,915 annotated material segments in 10,422 images spanning 22 different classes.

Most scale variations in KTHTIPS2b and OS are small compared with SForrest and ESVaT.  SForrest and ESVaT were specifically designed to test texture classification under extreme scale varyations.  However we do continue to test the performance of our framework on KTHTIPS2b and OS to show that the proposed can give significantly improved performance, even though our method is specifically designed for extreme scale variations.

For SForrest, half of the class samples were selected at random for training and the remaining half for testing, and results are reported over ten random partitionings of training and testing sets. For ESVaT, we split images evenly into training, validation and testing subsets. For KTHTIPS2b, one sample is available for training and the remaining three for testing, following \cite{Re6Cimpoi15}. For OS, we also use the same setup as in \cite{Re6Cimpoi15}.
SForrest and ESVaT are regrouped per Algorithm 1.  KTHTIPS2b and OS are augmented by building the image pyramids using the scale proposals as discussed in Sections~\ref{Sec:ScalePro}, but without regrouping.

\textbf{Implementation Details.}
We finetune VGGNet using CUReT \cite{Re7Dana1999} and UIUC \cite{Re15Lazebnik05}. Our finetuning is carried out for the whole network. The original training set of CUReT and UIUC are expanded as follows. For each image \textbf{I} in the training set $\Phi=\{\textbf{I}\}$, we compute its reduced scale proposal set $\mathbb{S}_{\textrm{sp}\textrm{-}\textrm{re}}$, then downsize it to obtain $M=|\mathbb{S}_{\textrm{sp}\textrm{-}\textrm{re}}|$ downsized images $\mathbb{I}^+=\{\textbf{I}_p\}$ using its scale proposals $\mathbb{S}_{\textrm{sp}\textrm{-}\textrm{re}}$. Thus, we have a new training set $\Phi^+=\mathbb{I}^+\bigcup\Phi$. Note that each image in $\mathbb{I}^+$ has the same class label as \textbf{I}.
We further perform data augmentation and window cropping on the new training set, following the method in \cite{Re28Simonyan2014VGG}. By these means, the number of training samples increases significantly, to 3.96 million, which are split at random into three even subsets ($\mathbb{S}_1$, $\mathbb{S}_2$ and $\mathbb{S}_3$), which are then used to train the three CNN models shown in Fig.~\ref{Fig:GANet} respectively.

In our GANet, we train three CNNs to perform crossover and mutation for the unit strings from the same layer because they show similar semantic features.  Since we propose to use three CNNs, we have three SVM classifiers. We use these three classifiers to vote for texture categories. Following the work in \cite{Re6Cimpoi15}, we also normalize descriptors by $L2$ norm and let the learning constant be $C=1$.

\begin {figure}[!t]
\centering
\includegraphics[width=0.45\textwidth]{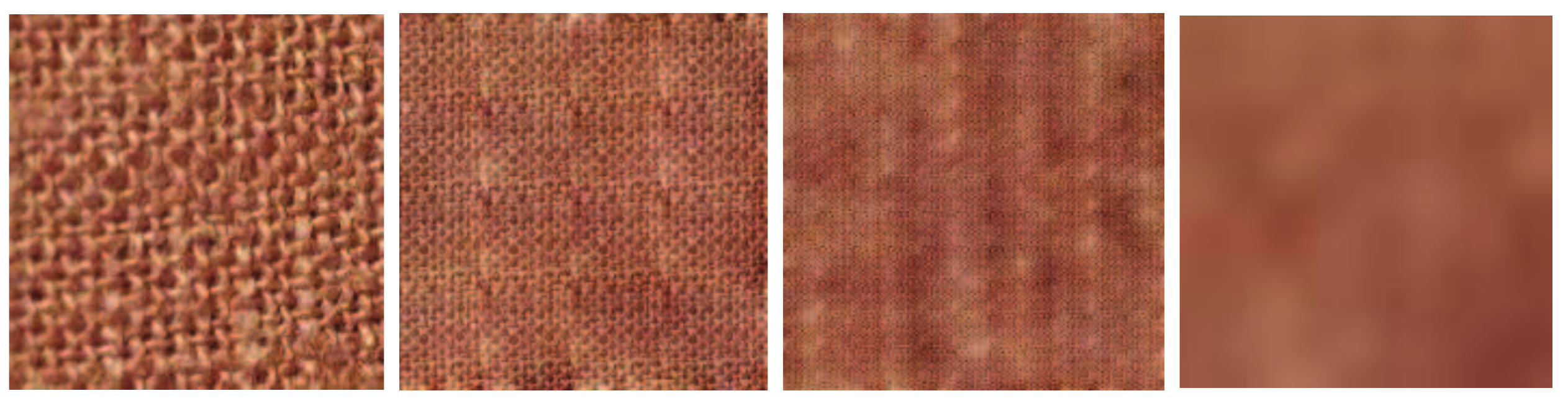}
\caption{Example of a synthesized texture at different scales.}
\label{Fig:Figure5}
\end {figure}

\subsection{Experimental Tests}
\label{Sec:Results}

\begin{table}[!t]
\caption {Classification scores of each CNN in GANet on the synthesized SForrest Dataset.}\label{Tab:EachCNN}
\centering
\renewcommand{\arraystretch}{1}
\setlength\tabcolsep{2pt}
\setlength\arrayrulewidth{0.1mm}
\begin{tabular}{!{\vrule width1.2bp}l|c!{\vrule width1.2bp}}
\Xhline{1pt}
\footnotesize Methods & \footnotesize SForrest \\
\Xhline{1pt}
\footnotesize FV-CNN ($\mathbb{S}_1$) & \footnotesize $85.3\%$ \\
\hline
\footnotesize FV-CNN ($\mathbb{S}_1+\mathbb{S}_2+\mathbb{S}_3$) & \footnotesize $87.3\%$ \\
\hline
\footnotesize FV-CNN-1 ($\mathbb{S}_1$) & \footnotesize $88.1\%$ \\
\hline
\footnotesize FV-CNN-2 ($\mathbb{S}_1$) & \footnotesize $87.8\%$ \\
\hline
\footnotesize FV-CNN-3 ($\mathbb{S}_2$) & \footnotesize $88.3\%$ \\
\hline
\footnotesize FV-GANet & \footnotesize $\textbf{90.4}\%$ \\
\Xhline{1pt}
\end{tabular}
\end{table}

\begin{table}[!t]
\caption {Performance evaluation of each component of our proposed scale searcher on the synthesized SForrest Dataset.}\label{Tab:GAmethod}
\centering
\renewcommand{\arraystretch}{1}
\setlength\tabcolsep{2pt}
\setlength\arrayrulewidth{0.1mm}
\begin{tabular}{!{\vrule width1.2bp}c|c|c|c!{\vrule width1.2bp}}
\Xhline{1pt}
\footnotesize Dataset &\multicolumn{3}{c!{\vrule width1.2bp}}{ \footnotesize SForrest }\\
\hline
\footnotesize Methods & \footnotesize Crossover& \footnotesize Mutation & \footnotesize Crossover+Mutation  \\
\Xhline{1pt}
\footnotesize FV-CNN-1 ($\mathbb{S}_1$) & \footnotesize $87.3\%$& \footnotesize $87.1\%$& \footnotesize $88.1\%$ \\
\hline
\footnotesize FV-CNN-2 ($\mathbb{S}_2$) & \footnotesize $87.1\%$& \footnotesize $86.8\%$& \footnotesize $87.8\%$ \\
\hline
\footnotesize FV-CNN-3 ($\mathbb{S}_3$) & \footnotesize $87.5\%$& \footnotesize $87.4\%$& \footnotesize $88.3\%$ \\
\hline
\Xhline{1pt}
\end{tabular}
\end{table}

\begin{table}[!t]
\caption {Performance evaluation of each component of our proposed framework on SForrest and KTHTIPS2b.}\label{Tab:GAmethod2}
\centering
\renewcommand{\arraystretch}{1}
\setlength\tabcolsep{2pt}
\setlength\arrayrulewidth{0.1mm}
\begin{tabular}{!{\vrule width1.2bp}l|c|c!{\vrule width1.2bp}}
\Xhline{1pt}
\footnotesize Methods & \footnotesize SForrest & \footnotesize KTHTIPS2b \\
\Xhline{1pt}
\footnotesize FV-GANet & \footnotesize $90.4\%$ & \footnotesize $82.6\%$ \\
\hline
\footnotesize FV-GANet+SP & \footnotesize $91.5\%$ & \footnotesize $83.0\%$ \\
\hline
\footnotesize FV-GANet+SP+RE& \footnotesize $96.3\%$ & \footnotesize $86.7\%$ \\
\Xhline{1pt}
\end{tabular}
\end{table}

\begin{table}[!t]
\caption {Performance evaluation using different scale proposal searchers on SForrest and KTHTIPS2b.}\label{Tab:GAmethod3}
\centering
\renewcommand{\arraystretch}{1}
\setlength\tabcolsep{2pt}
\setlength\arrayrulewidth{0.1mm}
\begin{tabular}{!{\vrule width1.2bp}l|c|c!{\vrule width1.2bp}}
\Xhline{1pt}
\footnotesize Methods & \footnotesize SForrest & \footnotesize KTHTIPS2b \\
\Xhline{1pt}
\footnotesize FV-GANet & \footnotesize $90.4\%$ & \footnotesize $82.6\%$ \\
\hline
\footnotesize FV-GANet+FFT & \footnotesize $89.8\%$ & \footnotesize $81.6\%$ \\
\hline
\footnotesize FV-GANet+Lindeberg& \footnotesize $92.1\%$ & \footnotesize $83.3\%$ \\
\hline
\footnotesize FV-GANet+SP+RE& \footnotesize $96.3\%$ & \footnotesize $86.7\%$ \\
\Xhline{1pt}
\end{tabular}
\end{table}

\textbf{GANet:} The classification results on SForrest are listed in Table~\ref{Tab:EachCNN}. FV-CNN means  the original VGGNet. FV-CNN ($\mathbb{S}_1$) means that FV-CNN is finetuned using $\mathbb{S}_1$. FV-CNN ($\mathbb{S}_1+\mathbb{S}_2+\mathbb{S}_3$) means that FV-CNN is finetuned using all the three subset $\mathbb{S}_1+\mathbb{S}_2+\mathbb{S}_3$. FV-CNN-\emph{n} ($n=1,2,3$) are the three CNN models finetuned using GA. FV-GANet is to combine the three FV-CNN-\emph{n} models by voting for texture categories.

From Table~\ref{Tab:EachCNN}, we can observe that including GA improves performance, since the individual models FV-CNN-\emph{n} outperform both FV-CNN ($\mathbb{S}_1$) and FV-CNN ($\mathbb{S}_1+\mathbb{S}_2+\mathbb{S}_3$). When combined, one can observe that FV-GANet is significantly better than FV-CNN.

The performance of using only crossover or mutation is shown in Table~\ref{Tab:GAmethod}. We used three subsets to train three CNNs separately. From this table, we can observe that crossover works slightly better than mutation. One reason might be that we get two new children strings for crossover but only one new child string for mutation. Thus, the former brings more diversity into the network.

\textbf{Component of Scale Searcher:} Scale searcher results are shown in Table~\ref{Tab:GAmethod2}. FV-GANet means that we only use FV-GANet. FV-GANet+SP means we use FV-GANet and the Scale Proposals (SP) component. FV-GANet+SP+RE means we use FV-GANet and the REduced SP.

The results in Table~\ref{Tab:GAmethod2} clearly show that the combination of FV-GANet, SP and RE improves the classification performance significantly. It demonstrates that texture classification in extreme scale variations can benefit from SP. After we combine RE, the number of scale proposals drops significantly since RE discards many errors in scale proposals.  To check the accuracy of the predicted scale levels for each image by FV-GANet+SP+RE, we use the mode of the predicted scale levels to compare with ground truth; the accuracy for the synthesized dataset is $97.4\%$, clearly demonstrating the accuracy of  texture classification in the extreme scales of SForrest.

\textbf{Different SP Searchers.} We compare our SP method with other possible SP searchers such as Fast Fourier Transform (FFT) and the method by Lindeberg \cite{Re17Lindeberg1998}. Results are shown in Table~\ref{Tab:GAmethod3}, which clearly demonstrate that our method works the best.
One possible reason that FV-GANet+FFT works poorly is that the texture images in the test set have other uninformative variations, such as illumination and rotation changes, besides  scale variations.

\begin {figure}[!t]
\centering
\includegraphics[width=0.3\textwidth]{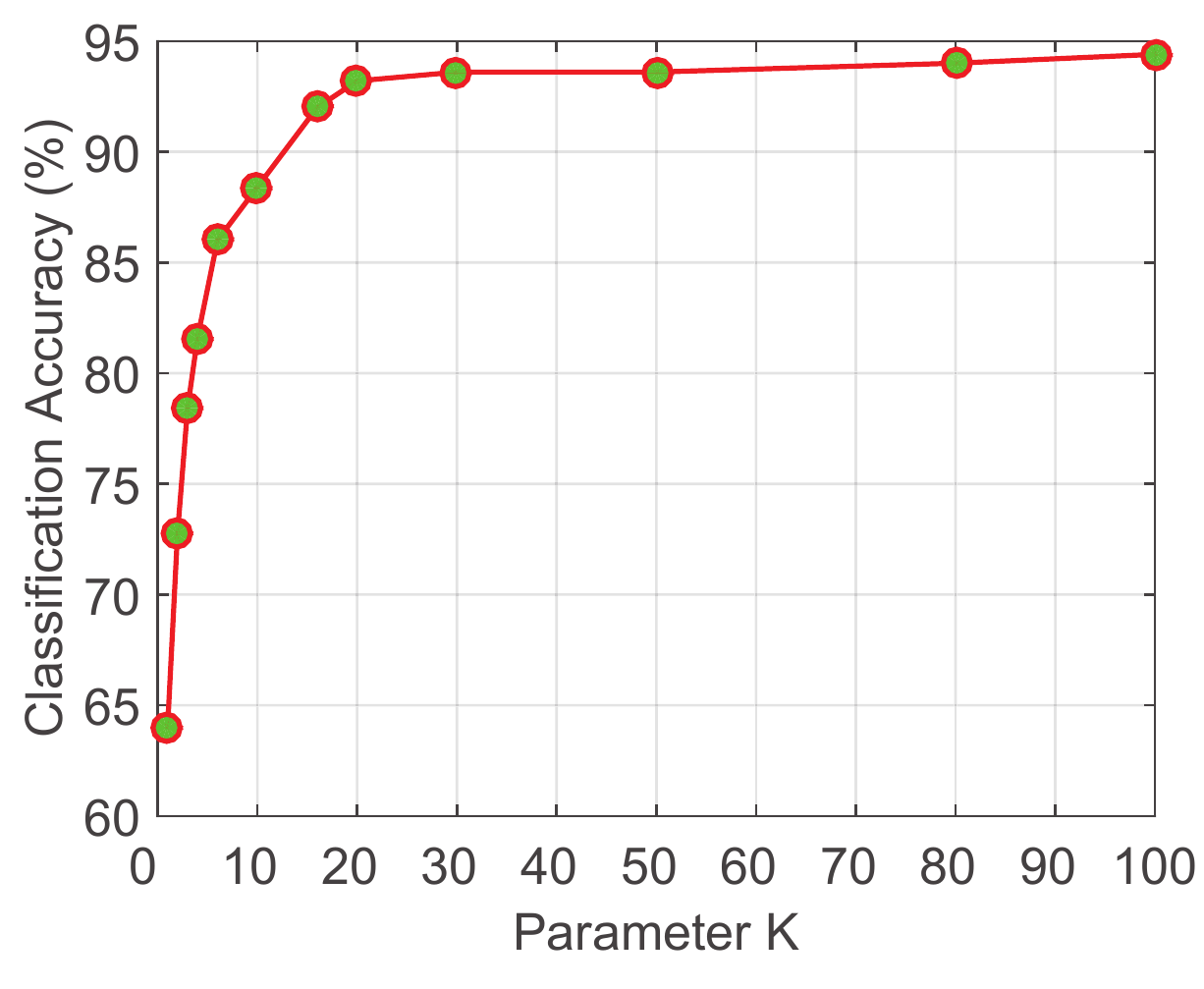}
\caption{Classification performance as a function of paramter $K$.}
\label{Fig:Figure6}
\end {figure}

\begin{table}[!t]
\caption {A comparison of our proposed method with the state-of-the-art in texture descriptors.}\label{Tab:Compare}
\centering
\renewcommand{\arraystretch}{1}
\setlength\tabcolsep{2pt}
\setlength\arrayrulewidth{0.1mm}
\begin{tabular}{!{\vrule width1.2bp}c|c|c|c|c!{\vrule width1.2bp}}
\Xhline{1pt}
\footnotesize Methods & \footnotesize SForrest& \footnotesize KTHTIPS2b & \footnotesize OS & \footnotesize ESVaT  \\
\Xhline{1pt}
\footnotesize LBP \cite{Re22Ojala02} & \footnotesize $61.8\%$& \footnotesize $49.9\%$& \footnotesize $35.4\%$& \footnotesize $40.5\%$ \\
\hline
\footnotesize SIFT \cite{Re19Lowe2004} & \footnotesize $63.4\%$& \footnotesize $52.7\%$& \footnotesize $38.9\%$& \footnotesize $39.1\%$ \\
\hline
\footnotesize IFV+DeCAF \cite{Re5Cimpoi2014} & \footnotesize $78.4\%$& \footnotesize $77.5\%$& \footnotesize $55.6\%$& \footnotesize $53.3\%$ \\
\hline
\footnotesize PR-proj \cite{Re27Simonyan2014} & \footnotesize $73.7\%$& \footnotesize $72.6\%$& \footnotesize $53.8\%$& \footnotesize $52.7\%$ \\
\hline
\footnotesize RAID-G \cite{Re35Wang2016} & \footnotesize N/A & \footnotesize $81.3\%$& \footnotesize $61.1\%$& \footnotesize N/A \\
\hline
\footnotesize FV-CNN & \footnotesize $85.3\%$& \footnotesize $82.1\%$& \footnotesize $60.3\%$& \footnotesize $56.2\%$ \\
\hline
\footnotesize LBP ($\mathbb{S}_1$)& \footnotesize N/A & \footnotesize $81.3\%$& \footnotesize N/A& \footnotesize N/A \\
\hline
\footnotesize B-CNN \cite{Re14Lamos2012} & \footnotesize N/A & \footnotesize $79.0\%$& \footnotesize N/A& \footnotesize N/A \\
\hline
\footnotesize Ours & \footnotesize $\textbf{96.3}\%$& \footnotesize $\textbf{86.7}\%$& \footnotesize $\textbf{68.4}\%$& \footnotesize $\textbf{67.9}\%$ \\
\Xhline{1pt}
\end{tabular}
\end{table}

\textbf{Parameter Evaluation:} In our approach, we have two important threshold parameters: $\eta$, the threshold similarity measure of two texels, and $K$, the number of similar texels of a candidate texel, as discussed in Section~\ref{Sec:ScalePro}. For parameter $\eta$, we use the following statistical value. Starting with dataset CUReT having 61 texture classes $\mathbb{S}=\{\mathbb{S}_1,...,\mathbb{S}_C\}$, we adopt LBP and nearest neighbor to  classify the textures in CUReT. For those correctly classified textures in each class we compute the image-wise similarity based on LBP histograms, letting $\phi_i$ denote the minimal similarity in class $\mathbb{S}_i$, and then setting $\eta=\min\{\phi_i\}$. We find that although $\eta$ is derived based on CUReT it works well for other datasets.

For parameter $K$, we determine its value empirically as shown in Fig.~\ref{Fig:Figure6}. The performance (and computational complexity) of our method increases with $K$, with performance very much levelling off at $K=20$, so we have set $K=20$ for all experiments.

\textbf{Results for KTHTIPS2b:} The results for KTHTIPS2b are shown in Table~\ref{Tab:GAmethod3}. One can observe that the performance improvement for KTHTIPS2b is similar to that for the SForrest. This demonstrates that our scale proposal approach generalizes to improving the texture classification in KTHTIPS2b which has small scale variations. Note that although images in KTHTIPS2b are spread over nine scales, we can find that there also several scale variations within one image.

\section{Comparative Evaluation \& Conclusions}

For comparison purposes, Table~\ref{Tab:Compare} compares our proposed approach with recent state of the art texture features including LBP \cite{Re22Ojala02}, SIFT \cite{Re19Lowe2004}, IFV+DeCAF \cite{Re5Cimpoi2014}, PR-proj \cite{Re27Simonyan2014}, RAID-G \cite{Re35Wang2016} and B-CNN \cite{Re14Lamos2012}.  For methods IFV+DeCAF, PR-proj and FV-CNN, we adopt the code provided by the original authors. For LBP and SIFT we use our own implementation. Note that IFV+DeCAF, PR-proj and FV-CNN are all fine-tuned using the same training set as our approach. The results of RAID-G \cite{Re35Wang2016} and B-CNN \cite{Re14Lamos2012} are quoted directly from their original papers.

From Table~\ref{Tab:Compare}, we can see that our proposed approach achieves consistently and significantly better results than all methods in comparison on all four evaluated datasets, particularly providing a significant performance margin in excess of 10\% on SForrest and ESVaT over previous state of the art, almost certainly because of the extreme scale variations present in those datasets, relative to the more modest gains in KTHTIPS2b, which has less extensive scale variations.

Overall, we have proposed a highly effective framework for recognizing textures with extreme scale variations, first searching scale proposals and then discarding errors in scale proposals by exploring dominant texture primitives. We have proposed a novel GANet network for better texture feature learning. Extensive experiments on four challenging texture benchmarks show that the proposed framework works much better than existing methods, especially for those textures with extreme scale variations. In addition, our method is  efficient since the scale proposal searching and reducing methods are very fast.
\bibliographystyle{IEEEtran}
\bibliography{IEEEabrv,egbib}

\begin{thebibliography}{10}
\providecommand{\url}[1]{#1}
\csname url@samestyle\endcsname
\providecommand{\newblock}{\relax}
\providecommand{\bibinfo}[2]{#2}
\providecommand{\BIBentrySTDinterwordspacing}{\spaceskip=0pt\relax}
\providecommand{\BIBentryALTinterwordstretchfactor}{4}
\providecommand{\BIBentryALTinterwordspacing}{\spaceskip=\fontdimen2\font plus
\BIBentryALTinterwordstretchfactor\fontdimen3\font minus
  \fontdimen4\font\relax}
\providecommand{\BIBforeignlanguage}[2]{{%
\expandafter\ifx\csname l@#1\endcsname\relax
\typeout{** WARNING: IEEEtran.bst: No hyphenation pattern has been}%
\typeout{** loaded for the language `#1'. Using the pattern for}%
\typeout{** the default language instead.}%
\else
\language=\csname l@#1\endcsname
\fi
#2}}
\providecommand{\BIBdecl}{\relax}
\BIBdecl

\bibitem{Pietikainen11}
M.~Pietik\"{a}inen, A.~Hadid, G.~Zhao, and T.~Ahonen,
  \emph{\BIBforeignlanguage{English}{Computer vision using local binary
  patterns}}.\hskip 1em plus 0.5em minus 0.4em\relax London, UK: Springer,
  2011.

\bibitem{Re15Lazebnik05}
S.~Lazebnik, C.~Schmid, and J.~Ponce, ``\BIBforeignlanguage{English}{A sparse
  texture representation using local affine regions},''
  \emph{\BIBforeignlanguage{English}{IEEE TPAMI}}, vol.~27, no.~8, pp.
  1265--1278, 2005.

\bibitem{liu2017local}
L.~Liu, P.~Fieguth, Y.~Guo, X.~Wang, and M.~Pietik{\"a}inen,
  ``\BIBforeignlanguage{English}{Local binary features for texture
  classification: Taxonomy and experimental study},''
  \emph{\BIBforeignlanguage{English}{Pattern Recognition}}, vol.~62, pp.
  135--160, 2017.

\bibitem{Re22Ojala02}
T.~Ojala, M.~Pietik\"{a}inen, and T.~M\"{a}enp\"{a}\"{a},
  ``\BIBforeignlanguage{English}{Multiresolution gray-scale and rotation
  invariant texture classification with local binary patterns},''
  \emph{\BIBforeignlanguage{English}{IEEE TPAMI}}, vol.~24, no.~7, pp.
  971--987, July 2002.

\bibitem{Re26Shechtman2007}
E.~Shechtman and M.~Irani, ``\BIBforeignlanguage{English}{Matching local
  self-similarities across images and videos},'' in
  \emph{\BIBforeignlanguage{English}{CVPR}}, 2007, pp. 1--8.

\bibitem{Re5Cimpoi2014}
M.~Cimpoi, S.~Maji, I.~Kokkinos, S.~Mohamed, and A.~Vedaldi,
  ``\BIBforeignlanguage{English}{Describing textures in the wild},'' in
  \emph{\BIBforeignlanguage{English}{CVPR}}, 2014, pp. 3606--3613.

\bibitem{Re18Liu2016}
L.~Liu, P.~Fieguth, X.~Wang, M.~Pietik{\"a}inen, and D.~Hu,
  ``\BIBforeignlanguage{English}{Evaluation of lbp and deep texture descriptors
  with a new robustness benchmark},'' in
  \emph{\BIBforeignlanguage{English}{ECCV}}, 2016, pp. 69--86.

\bibitem{Re35Wang2016}
Q.~Wang, P.~Li, W.~Zuo, and L.~Zhang, ``\BIBforeignlanguage{English}{{RAID-G}:
  Robust estimation of approximate infinite dimensional gaussian with
  application to material recognition},'' in
  \emph{\BIBforeignlanguage{English}{CVPR}}, 2016, pp. 4433--4441.

\bibitem{Liu2016Median}
L.~Liu, S.~Lao, P.~Fieguth, Y.~Guo, X.~Wang, and M.~Pietik{\"a}inen,
  ``\BIBforeignlanguage{English}{Median robust extended local binary pattern
  for texture classification},'' \emph{\BIBforeignlanguage{English}{IEEE TIP}},
  vol.~25, no.~3, pp. 1368--1381, 2016.

\bibitem{Xu10}
Y.~Xu, X.~Yang, H.~Ling, and H.~Ji, ``\BIBforeignlanguage{English}{A new
  texture descriptor using multifractal analysis in multiorientation wavelet
  pyramid},'' in \emph{\BIBforeignlanguage{English}{CVPR}}, 2010, pp. 161--168.

\bibitem{Re6Cimpoi15}
M.~Cimpoi, S.~Maji, and A.~Vedaldi, ``\BIBforeignlanguage{English}{Deep filter
  banks for texture recognition and segmentation},'' in
  \emph{\BIBforeignlanguage{English}{CVPR}}, 2015, pp. 3828--3836.

\bibitem{Re2Brodatz66}
P.~Brodatz, \emph{\BIBforeignlanguage{English}{Textures: A Photographic Album
  for Artists and Designers}}.\hskip 1em plus 0.5em minus 0.4em\relax New York,
  USA: Dover Publications, 1966.

\bibitem{Re7Dana1999}
K.~Dana, B.~V. Ginneken, S.~Nayar, and J.~Koenderink,
  ``\BIBforeignlanguage{English}{Reflectance and texture of real-world
  surfaces},'' \emph{\BIBforeignlanguage{English}{ACM TOG}}, vol.~18, no.~1,
  pp. 1--34, 1999.

\bibitem{Re3Caputo05}
B.~Caputo, E.~Hayman, and P.~Mallikarjuna,
  ``\BIBforeignlanguage{English}{Class-specific material categorization},'' in
  \emph{\BIBforeignlanguage{English}{ICCV}}, 2005, pp. 1597--1604.

\bibitem{Re1Bell2013}
S.~Bell, P.~Upchurch, N.~Snavely, and K.~Bala,
  ``\BIBforeignlanguage{English}{Opensurfaces: A richly annotated catalog of
  surface appearance},'' \emph{\BIBforeignlanguage{English}{ACM TOG}}, vol.~32,
  no.~4, p. 111, 2013.

\bibitem{LiuFieguthPAMI}
L.~Liu and P.~Fieguth, ``\BIBforeignlanguage{English}{Texture classification
  from random features},'' \emph{\BIBforeignlanguage{English}{IEEE TPAMI}},
  vol.~34, no.~3, pp. 574 --586, March 2012.

\bibitem{Zhang07}
J.~Zhang, M.~Marszalek, S.~Lazebnik, and C.~Schmid,
  ``\BIBforeignlanguage{English}{Local features and kernels for classification
  of texture and object categories: a comprehensive study},''
  \emph{\BIBforeignlanguage{English}{IJCV}}, vol.~73, no.~2, pp. 213--238,
  2007.

\bibitem{Julesz1981}
B.~Julesz, ``\BIBforeignlanguage{English}{Textons, the elements of texture
  perception, and their interactions},''
  \emph{\BIBforeignlanguage{English}{Nature}}, vol. 290, no. 5802, pp. 91--97,
  1981.

\bibitem{Re33Xie08}
X.~Xie and M.~Mirmehdi, \emph{\BIBforeignlanguage{English}{A galaxy of texture
  features--Handbook of texture analysis}}.\hskip 1em plus 0.5em minus
  0.4em\relax London, UK: Imperial College Press, 2008.

\bibitem{Re17Lindeberg1998}
T.~Lindeberg, ``\BIBforeignlanguage{English}{Feature detection with automatic
  scale selection},'' \emph{\BIBforeignlanguage{English}{IJCV}}, vol.~30,
  no.~2, pp. 79--116, 1998.

\bibitem{Re21Mirmehdi2000}
M.~Mirmehdi and M.~Petrou, ``\BIBforeignlanguage{English}{Segmentation of color
  textures},'' \emph{\BIBforeignlanguage{English}{IEEE TPAMI}}, vol.~22, no.~2,
  pp. 142--159, 2000.

\bibitem{Re4Cheng2014bing}
M.~Cheng, Z.~Zhang, W.~Lin, and P.~Torr, ``\BIBforeignlanguage{English}{{BING}:
  Binarized normed gradients for objectness estimation at 300fps},'' in
  \emph{\BIBforeignlanguage{English}{CVPR}}, 2014, pp. 3286--3293.

\bibitem{Re10He2016ResNet}
K.~He, X.~Zhang, S.~Ren, and J.~Sun, ``\BIBforeignlanguage{English}{Deep
  residual learning for image recognition},'' in
  \emph{\BIBforeignlanguage{English}{CVPR}}, 2016, pp. 770--778.

\bibitem{Re28Simonyan2014VGG}
K.~Simonyan and A.~Zisserman, ``\BIBforeignlanguage{English}{Very deep
  convolutional networks for large-scale image recognition},'' in
  \emph{\BIBforeignlanguage{English}{ICLR}}, 2015.

\bibitem{Re13AlexNet2012}
A.~Krizhevsky, I.~Sutskever, and G.~Hinton,
  ``\BIBforeignlanguage{English}{Imagenet classification with deep
  convolutional neural networks},'' in
  \emph{\BIBforeignlanguage{English}{NIPS}}, 2012, pp. 1097--1105.

\bibitem{Re27Simonyan2014}
K.~Simonyan, A.~Vedaldi, and A.~Zisserman,
  ``\BIBforeignlanguage{English}{Learning local feature descriptors using
  convex optimisation},'' \emph{\BIBforeignlanguage{English}{IEEE TPAMI}},
  vol.~36, no.~8, pp. 1573--1585, 2014.

\bibitem{Re38Zhoubolei2014}
B.~Zhou, A.~Khosla, A.~Lapedriza, A.~Oliva, and A.~Torralba,
  ``\BIBforeignlanguage{English}{Object detectors emerge in deep scene cnns},''
  in \emph{\BIBforeignlanguage{English}{ICLR}}, 2014.

\bibitem{Re8David2014}
O.~David and I.~Greental, ``\BIBforeignlanguage{English}{Genetic algorithms for
  evolving deep neural networks},'' in
  \emph{\BIBforeignlanguage{English}{Annual Conference on Genetic and
  Evolutionary Computation}}, 2014, pp. 1451--1452.

\bibitem{Re14Lamos2012}
J.~Lamos-Sweeney, ``\BIBforeignlanguage{English}{Deep learning using genetic
  algorithms},'' \emph{\BIBforeignlanguage{English}{Thesis of Rochester
  Institute of Technology}}, 2012.

\bibitem{Re29Steininger2016}
C.~Steininger, ``\BIBforeignlanguage{English}{Genetic algorithms with deep
  learning for robot navigation},''
  \emph{\BIBforeignlanguage{English}{Thesis}}, 2016.

\bibitem{Xie2017ICCV}
L.~Xie and A.~Yuille, ``\BIBforeignlanguage{English}{Genetic {CNN}},'' in
  \emph{\BIBforeignlanguage{English}{ICCV}}, 2017.

\bibitem{Ding2013}
S.~Ding, H.~Li, C.~Su, J.~Yu, and F.~Jin,
  ``\BIBforeignlanguage{English}{Evolutionary artificial neural networks: A
  review},'' \emph{\BIBforeignlanguage{English}{Artificial Intelligence
  Review}}, vol.~39, no.~3, pp. 251--260, 2013.

\bibitem{Webb2011}
A.~Webb and K.~Copsey, \emph{Statistical Pattern Recognition (Third
  Edition)}.\hskip 1em plus 0.5em minus 0.4em\relax Wiley, 2011.

\bibitem{Re11hinton2006}
G.~Hinton and R.~Salakhutdinov, ``\BIBforeignlanguage{English}{Reducing the
  dimensionality of data with neural networks},''
  \emph{\BIBforeignlanguage{English}{science}}, vol. 313, no. 5786, pp.
  504--507, 2006.

\bibitem{Re23Olshausen1996}
B.~Olshausen and D.~Field, ``\BIBforeignlanguage{English}{Emergence of
  simple-cell receptive field properties by learning a sparse code for natural
  images},'' \emph{\BIBforeignlanguage{English}{Nature}}, vol. 381, no. 6583,
  p. 607, 1996.

\bibitem{Re39Forrest}
``\BIBforeignlanguage{English}{The forrest dataset},''
  \url{http://textures.forrest.cz/}.

\bibitem{Re25Perez2003}
P.~P{\'e}rez, M.~Gangnet, and A.~Blake, ``\BIBforeignlanguage{English}{Poisson
  image editing},'' in \emph{\BIBforeignlanguage{English}{ACM TOG}}, vol.~22,
  no.~3, 2003, pp. 313--318.

\bibitem{Re19Lowe2004}
D.~Lowe, ``\BIBforeignlanguage{English}{Distinctive image features from
  scale-invariant keypoints},'' \emph{\BIBforeignlanguage{English}{IJCV}},
  vol.~60, no.~2, pp. 91--110, 2004.

\end{thebibliography}

\end{document}